%% file: main.tex
\documentclass[conference]{IEEEtran}
\IEEEoverridecommandlockouts
\usepackage{cite}

\usepackage{amsmath,amssymb,amsfonts}
\usepackage{graphicx}
\usepackage{textcomp}
\usepackage{xcolor}
\usepackage{hyperref}
\usepackage{caption}
\usepackage{subcaption}
\usepackage{algorithm}
\usepackage{algpseudocode}
\usepackage{tabulary}
\usepackage{multirow}
\def\BibTeX{{\rm B\kern-.05em{\sc i\kern-.025em b}\kern-.08em
    T\kern-.1667em\lower.7ex\hbox{E}\kern-.125emX}}
\usepackage{acro}[=v2]
\input{acronyms}

\makeatletter 
\newcommand{\linebreakand}{%
  \end{@IEEEauthorhalign}
  \hfill\mbox{}\par
  \mbox{}\hfill\begin{@IEEEauthorhalign}
}
\makeatother 

\begin{document}

\title{Offline Skill Generalization via Task and Motion Planning
}

\author{\IEEEauthorblockN{Shin Watanabe*}
\and
\IEEEauthorblockN{Geir Horn}
\and
\IEEEauthorblockN{Jim Tørresen}
\and
\IEEEauthorblockN{Kai Olav Ellefsen}
\linebreakand
\IEEEauthorblockA{\textit{Department of Informatics} \\
\textit{University of Oslo} \\
Oslo, Norway}
}

\maketitle

\input{abstract}

{\footnotesize \textsuperscript{*}Corresponding author: shinwa@ifi.uio.no}
\input{introduction} 

\input{related_work}

\input{background}

\input{method}

\input{experiments}

\input{results}

\input{conclusion}

\bibliographystyle{IEEEtran}
\bibliography{references, extra}


\end{document}

%% file: acronyms.tex
\usepackage{acro}[=v2]

\DeclareAcronym{TAMP}{
  short = TAMP,
  long  = Task-and-motion planning,
  class = abbrev
}

\DeclareAcronym{RL}{
  short = RL,
  long  = Reinforcement Learning,
  class = abbrev
}

\DeclareAcronym{IL}{
  short = IL,
  long  = Imitation Learning,
  class = abbrev
}

\DeclareAcronym{LfD}{
  short = LfD,
  long  = Learning from Demonstrations,
  class = abbrev
}

\DeclareAcronym{PDDL}{
  short = PDDL,
  long  = Planning Domain Definition Language,
  cite = {aeronautiques1998pddl},
  class = abbrev
}

\DeclareAcronym{IK}{
  short = IK,
  long  = Inverse Kinematics,
  class = abbrev
}

\DeclareAcronym{DoF}{
	short = DoF,
	long  = Degree-of-Freedom,
	class = abbrev
}

\DeclareAcronym{BC}{
	short = BC,
	long  = Behavioral Cloning,
	cite = {pomerleau_alvinn_1988},
	class = abbrev
}

\DeclareAcronym{ORL}{
 short = ORL,
 long = Offline RL,
 class = abbrev
}

\DeclareAcronym{BCQ}{
	short = BCQ,
	long  = Batch-constrained Q-learning,
	cite = {fujimoto_off-policy_2019},
	class = abbrev
}

%% file: abstract.tex
\begin{abstract}
This paper presents a novel approach to generalizing robot manipulation skills by combining a sampling-based task-and-motion planner with an offline reinforcement learning algorithm. Starting with a small library of scripted primitive skills (e.g. Push) and object-centric symbolic predicates (e.g. On(block, plate)), the planner autonomously generates a demonstration dataset of manipulation skills in the context of a long-horizon task. An offline reinforcement learning algorithm then extracts a policy from the dataset without further interactions with the environment and replaces the scripted skill in the existing library. Refining the skill library improves the robustness of the planner, which in turn facilitates data collection for more complex manipulation skills.

We validate our approach in simulation, on a block-pushing task.
We show that the proposed method requires less training data than conventional reinforcement learning methods.
Furthermore, interaction with the environment is collision-free because of the use of planner demonstrations, making the approach more amenable to persistent robot learning in the real world.
\end{abstract}


%% file: introduction.tex
\section{Introduction}

Generalizing robot manipulation skills to multiple environments in long-horizon tasks remains a grand challenge in robot learning, despite substantial progress in the recent decade \cite{openai_solving_2019, kalashnikov_mt-opt_2021}.
Much of this progress can be attributed to advancements in deep learning and their integration with online \ac{RL}, which now stands as the dominant approach to robot skill learning.
A recent review article \cite{ibarz_how_2021} lays out several outstanding hurdles in online \ac{RL} for robotics.
These include
\textit{sample efficiency} (reducing the amount of costly environment interactions),
\textit{robot persistence} (increasing the capability of robots to collect data with minimal human intervention), and
 \textit{safe learning} (avoiding self-collisions, jerky actuation, collisions with obstacles, etc.).
In addition, online \ac{RL} struggles to solve \textit{long-horizon} tasks  \cite{zhu_robosuite_2022}.

Concurrently to online \ac{RL}, several other approaches either have sought to address these challenges or are well-suited to bypass them entirely.
In particular, \ac{LfD} circumvents the need for exploration by leveraging demonstration data, which naturally provides a dense learning signal. \ac{LfD}-based approaches such as \ac{IL} and offline \ac{RL} have been shown to efficiently generalize manipulation skills on a variety of demonstration learning tasks \cite{mandlekar_what_2023}.

However, expert demonstrations are cumbersome or impractical to collect for long-horizon manipulation tasks. Additionally, \ac{LfD} can suffer from compounding errors, if the provided demonstrations do not cover the state space encountered during evaluation \cite{ross_reduction_2011}.
This motivates the use of an approach that is capable of providing diverse demonstration data for long-horizon manipulation tasks autonomously.

\ac{TAMP} is a well-established approach for robot control, where skills are first represented symbolically so that the abstract skill sequence to accomplish a given task is generated by a classical planner and motion trajectories specific to the environment instance are generated by a motion planner. \ac{TAMP} approaches excel at long-horizon tasks due to its strong action abstraction, and can generalize skills across tasks and domains due to its strong state abstraction \cite{garrett_integrated_2021}. In addition, the trajectory provided by the motion planner is guaranteed to be kinematically feasible and collision-free when provided with an accurate environment perception, making this approach an attractive option for safe learning.

Our work leverages complementary features of \ac{LfD} and \ac{TAMP} to introduce a skill learning architecture that is competitive with state-of-the-art online \ac{RL} methods while addressing the aforementioned challenges. Starting with a small set of engineered primitive skills, object-centric predicates, a symbolic description of the skill to be learned, and a scripted subroutine, our approach autonomously generalizes the skill across task and environment instances through demonstration data generated by a \ac{TAMP} solver (see Figure \ref{fig:overview} for a schematic).
Since the data collection is performed as an integral part of the larger long-horizon task, the demonstrations are contextually more relevant than if the skills were learned in isolation \cite{cheng_guided_2022}.
For example, a robot arm equipped with a mobile base learning to push an object should learn the skill in a variety of base configurations, because in a real-world setting it is not realistic to assume that the mobile base is always aligned with the object in the same way.
The result is a self-reinforcing cycle between \ac{TAMP} and \ac{LfD}, where TAMP provides demonstration data for skill generalization, and the learned skill policy improves the robustness of TAMP.

In summary, our contributions are as follows:
\begin{enumerate}
    \item we propose a skill-learning architecture that combines standard \ac{LfD} algorithms with an existing sampling-based integrated \ac{TAMP} solver to autonomously generalize manipulation skills safely and robustly.
    \item we implement and evaluate this architecture to show that the learned skill can a) outperform the suboptimal expert, b) outperform an online \ac{RL} method with a fraction of environment interactions, and c) be reused for long-horizon tasks.
\end{enumerate}

%% file: related_work.tex
\section{Related Work}

\subsection{Integrated Task and Motion Planning}

\ac{TAMP} research aims to solve challenging robot control problems by splitting the problems into a discrete symbolic component, for which the automated planning community developed methods to address high-dimensional domains \cite{ghallab_automated_2016}, and a continuous geometric component, for which the robotics community developed methods to address practical domains \cite{lavalle_planning_2006}.
Integrated \ac{TAMP} refers to a family of \ac{TAMP} algorithms that seeks to combine the two components in a way that minimizes the runtime of the complete planning procedure \cite{garrett_integrated_2021}.
Our approach builds on a specific class of integrated TAMP, in which the task planner solves the symbolic problem first, then the motion planner verifies the feasibility of the specific task instance in continuous space.

\subsection{Learning for TAMP}
Most \ac{TAMP} approaches assume perfect observability, actuation, and model of the environment (e.g. object shapes and kinematics), each of which pose a significant restriction in terms of real-world applicability. Our work aims to relax the assumption of a known environment model by learning a model-free policy from demonstrations provided by TAMP. Crucially, these demonstrations are suboptimal because the motion planner does not account for the stochasticity of the object dynamics. Other work learns the constraints of a parametrized skill that lead to successful executions of the planner \cite{wang_learning_2021}. Parametrizing a skill is often not trivial because it requires conceptually decomposing the skill into multiple parameters. On the other hand, a scripted subroutine only requires a single state-to-action mapping with no knowledge of the underlying parameter space.

Our work assumes a fully observable environment, in contrast to Manipulation from Zero Models (M0M) \cite{curtis_long-horizon_2022} that learns object affordances directly from perceptual data, and generates task plans using the object affordances as constraints. M0M is complementary to our work because the manipulation skills they use in their approach such as picking and placing are scripted and not learned.

\subsection{Guided Skill Learning and Generalization}

Our work is similar to that of McDonald \textit{et al.} \cite{mcdonald_guided_2021}, in which an optimization-based integrated TAMP framework is used to guide the policy learning of robot manipulation skills. Their online approach biases the demonstration trajectory to stay close to the policy being trained. Our fully offline approach does not require interactions between the demonstrator and the learner. 

Dalal \textit{et al.} \cite{dalal_imitating_2023} use a Transformer-based \cite{vaswani_attention_2017} architecture to train a visuomotor policy in an offline manner from \ac{TAMP}-provided demonstrations using behavioral cloning. The resulting policy maps observations directly to actions, thereby replacing both the task planner and the motion planner.
In contrast to these approaches, our objective is to maintain the task planning component while expanding the repertoire of separately learned skills, which allows for abstract skill composition.

Our approach is most closely related to LEAGUE \cite{cheng_guided_2022}, which uses a \ac{TAMP} framework to guide robotic manipulation skill learning and generalization. In contrast with our approach, LEAGUE uses an online \ac{RL} method (Soft Actor-Critic \cite{haarnoja_soft_2018}) to interactively learn the policy, thereby subject to the aforementioned hurdles.
In particular, experimental results show that LEAGUE takes millions of environment interactions to learn a single skill.
Synergistic-\ac{TAMP} \cite{liu_synergistic_2023} also takes an online \ac{RL}-based approach to learning a pushing skill, whereby the plan feasibility acts as a reward mechanism for the skill learner, potentially subject to sparse rewards. In contrast, our approach guarantees every trajectory has a reward through expert relabeling.

\subsection{Expert Relabeling}
Inspired by the success of Hindsight Experience Replay \cite{andrychowicz_hindsight_2017} in the off-policy \ac{RL} setting, Ding \textit{et al.} \cite{ding_goal-conditioned_2019} boost the sample efficiency of goal-conditioned imitation learning through \textit{expert relabeling}: given an expert demonstration, every two points along the trajectory can be considered an initial state and a goal state, under the assumption that the paths are approximately geodesic.\footnote{This intuitively means that actions are taken to reach the goal as quickly as possible.} Our approach applies this insight to learn a goal-conditioned policy but without additional environment interactions, as the demonstrations are given by a \ac{TAMP} solver.

\begin{figure*}
\centerline{\includegraphics[width=0.8\textwidth]{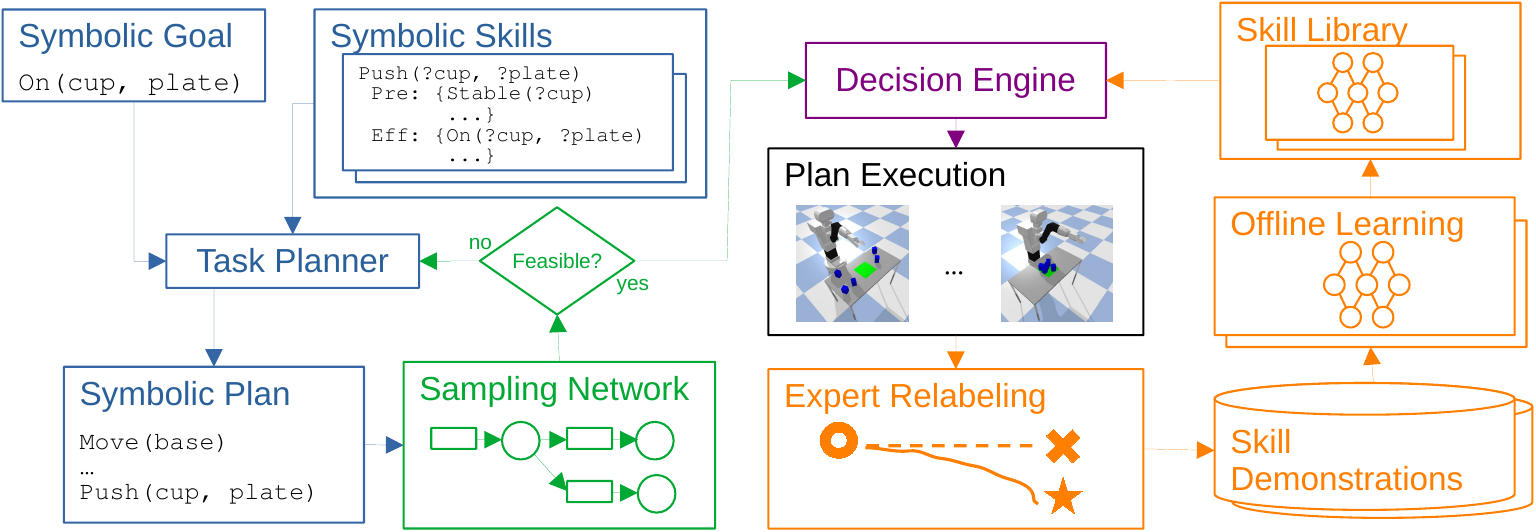}}
\caption{Method overview. Starting with symbolically specified skill operators, goals, and scripted subroutines (blue), a \ac{TAMP} solver generates a kinematically feasible and collision-free task plan (green), which is executed open-loop in the environment (black). By relabeling the goal of each suboptimal trajectory to its final state, a dataset with dense rewards is generated for each skill (orange). Given a sufficiently diverse dataset, an offline reinforcement learning algorithm can distill a reactive policy that outperforms the original subroutine and can therefore replace it for use in long-horizon tasks (purple).}
\label{fig:overview}
\end{figure*}

%% file: background.tex
\section{\label{background}Background}

\subsection{Task Planning}

A task planning problem can be formalized as a set of states \(\mathcal{S}\), a set of transitions \(\mathcal{T} \subseteq \mathcal{S} \times \mathcal{S}\), an initial state \(s_0 \in \mathcal{S}\), and a set of goal states \(\mathcal{S}_* \subseteq \mathcal{S}\). 
The objective for a planner is to find a plan \(\pi\), which is a sequence of transitions \(t = \langle s, s' \rangle \in \mathcal{T}\), that advances the initial state \(s_0\) into a goal state \(s_* \in \mathcal{S}_*\). 
To make the problem tractable, the state representation is factored into a collection of state variables \(\mathcal{V} = \{1, ... , m\}\), each of which has a finite domain \(\mathcal{X}_v\). 
Typically, these state variables correspond to properties of objects or relations among them in a domain (i.e. object-centric abstraction). 
In addition, transitions are described by a set of effects \texttt{eff} and preconditions \texttt{pre}: effects list state variables that are modified as a result of the transition, and preconditions list state variables under which the transition is allowed to occur.
A minimal example of a pushing skill specified in \ac{PDDL} is given in the blue box denoted \textit{Symbolic Skills} in Figure \ref{fig:overview}.
The \texttt{?} symbol prefixing the parameters denotes that they are not \textit{grounded} on an environment instance, i.e. they are \textit{lifted}.
The skill operator becomes \textit{grounded} when all the parameters are instantiated and all the predicates are evaluated on them.

\subsection{Task and Motion Planning}

The task planning formalism can be extended to handle geometric and kinematic constraints by way of \textit{streams} \cite{garrett_pddlstream_2020}, which generate instantiated predicates that support candidate task plans. For example, to execute the \texttt{push} skill introduced above, the skill must be kinematically and geometrically feasible, i.e. that the end-effector can reach the goal pose, and that there exists a collision-free trajectory from its initial pose. In this case, one stream calls the \ac{IK} solver to generate a valid joint configuration, and another stream calls a motion planner that generates collision-free trajectories. We refer readers to the original article \cite{garrett_pddlstream_2020} for a complete description of the language.

Unfortunately, satisfying these constraints are necessary but insufficient to execute arbitrary manipulation skills successfully, because assumptions about quasi-static, fully observable, deterministic environments do not generally hold.

\subsection{\label{sec:gcil}Goal-Conditioned Imitation Learning}

Our objective is to improve the performance, robustness and generalizability of \ac{TAMP}-generated manipulation skills through \ac{LfD}. In \ac{IL}, demonstrations are given in the form of observation-action trajectories \(\tau = \{o_0, a_0, o_1, a_1, ... , o_T, a_T\}\), and the goal is to find a mapping from observations to actions that imitates the demonstration. To mirror the action abstraction capabilities of a \ac{TAMP} framework, our distilled policy must not only be conditioned on initial states but also on goal states, so that the skill can reach any state configuration requested by the planner. In goal-conditioned behavioral cloning, the demonstration dataset  \(\{\tau^j\}^D_{j=0}\) is augmented with the goal observation to make observation-action-goal tuples  \((o^j_t, a^j_t, g^j)\) and supervised regression is run to obtain the policy \cite{ding_goal-conditioned_2019}.


Furthermore, even if the demonstration trajectory \(\tau^j\) does not reach the requested goal \(g^j\), the goal can be relabeled \textit{a posteriori} to any observation encountered in the transition \((o_t^j, a_t^j, g^{j'} = o_{t+k}^j)\), such that a suboptimal demonstration becomes optimal.

\subsection{Offline RL}

One limitation of \ac{IL}-based methods is that the learned policy can only perform as well as the (relabeled) demonstration, because they only seek to predict the expert's action given the current state. A framework that can potentially outperform the demonstration is offline \ac{RL}, where off-policy \ac{RL} algorithms are used to train policies entirely from the demonstration dataset \cite{kumar_should_2021}.
Recall that the goal of \ac{RL} is to learn a policy \(\pi(\cdot | s)\) that maximizes the expected cumulative discounted reward \(J(\pi)\) in a Markov decision process  \((\mathcal{S} , \mathcal{A}, P, r, \gamma)\). 
\(\mathcal{S} , \mathcal{A}\) represent state and action spaces,
\(P(\cdot | s, a)\) and \(r(s,a)\) represent the dynamics and mean reward function, and
\(\gamma \in (0, 1)\) represents the discount factor.
In the offline case, we are given a dataset of transitions \(\mathcal{D} = \{(s_i, a_i, r_i, s'_i)\}_{i=1}^N\) of size \(N\), which we assume is generated i.i.d. from a distribution \(\mu(s,a)\) that specifies the behavior policy \(\Pi_\beta(a|s)\), which is the subroutine in our case.
Note that the empirical dynamics and reward distributions \(\hat{P}(\cdot | s, a)\) and \(\hat{r}(s,a)\) seen in \(\mathcal{D}\) may be different from \(P\) and \(r\) due to stochasticity.
The challenge of offline \ac{RL} lies in overcoming the distribution shift, where the learned policy executes out-of-distribution actions.
The success of offline \ac{RL} depends on how well the data distribution \(\mu(s,a)\) covers the state-action pairs visited under the optimal policy.
We hypothesize that the coverage of state-action pairs by the demonstration dataset provided by the \ac{TAMP} solver enables the learned policy to remain in the distribution.

%% file: method.tex
\section{Methodology}

Below, we motivate and describe the complete skill-learning pipeline from obtaining the \ac{TAMP} demonstration trajectories to distilling a reactive goal-conditioned policy.
We consider a long-horizon manipulation task involving multiple objects, with a uniform distribution of initial states (i.e. poses). The \ac{TAMP} system is assumed to be equipped with a set of sampling-based skills, e.g. a base motion planner. The objective of our approach is to learn reactive policies that replace the scripted skills that have low success rates, described below.

\subsection{Scripting sub-optimal skills for TAMP}
We focus on learning a single policy per skill. Consider the scripted skill \texttt{Push}.
It consists of a skill operator description specified in \ac{PDDL} and several streams that take as input the initial configurations as well as geometric and kinematic constraints and output grounded predicates, such as feasible motion trajectories. 
We refer the readers to the linked code repository for the complete domain and stream specification.
The stream that outputs the arm trajectory producing the push motion assumes that the object is attached to the end-effector during the maneuver. Specifically, the arm trajectory consists of a joint interpolation between the initial end-effector pose behind the object such that the gripper side surface is orthogonal to the goal direction, and the goal end-effector pose such that the object would be at the goal pose if it was attached to the end-effector.
In effect, the motion planner does not take into account the rotation of the object as it is pushed, and since the trajectory is executed in an open-loop fashion, the object is not guaranteed to stay on the gripper surface as it gets pushed.

Note that successfully executing the push maneuver would require an accurate model of the environment dynamics (e.g. object shape, friction coefficients, etc.) which are generally unavailable or difficult to estimate, in addition to a closed-loop controller to handle the stochasticity.

\subsection{Collecting demonstrations from TAMP}
This work assumes access to ground-truth states directly from the simulator.
The observation space consists of the end-effector pose and the object pose,
where the rotation is expressed in quaternions,
resulting in a 14-dimensional vector.
The action space consists of the desired rotation change of every arm joint, resulting in a 7-dimensional vector.
The goal space consists of the object position in Cartesian coordinates.
The sampling frequency for the policy is 20 Hz.
All relevant states were sampled with respect to the robot base frame to account for the mobile base. 
To speed up the data collection procedure, the TAMP solver is given a computation budget of 30 seconds.

\subsection{Processing demonstrations for offline RL}
Following the prescriptions from Dalal \textit{et al.} \cite{dalal_imitating_2023}, demonstrations with unusually long trajectories were discarded to mitigate distribution shift. Specifically, a maximum horizon length was chosen to be 200.
Next, the goal pose given by the TAMP solver is replaced by the pose that the object ended up in after the subroutine execution (c.f. Section \ref{sec:gcil}).
Finally, the trajectory is pruned to the exact time step at which the object stops moving, to further contain the solution trajectory.

%% file: experiments.tex
\section{Experiment Setup}

We evaluate our approach on a goal-conditioned long-horizon manipulation task domain, in order to answer the following question:
Can automatically generated demonstrations from an integrated \ac{TAMP} framework be distilled into a reactive goal-conditioned policy that:
\begin{enumerate}
    \item improves upon the performance of the original demonstrations?
    \item outperforms an online goal-conditioned \ac{RL} policy with less data?
    \item generalizes to long-horizon tasks?
\end{enumerate}

\subsection{Environment Setup}

\begin{figure}
\centerline{\includegraphics[width=0.25\textwidth]{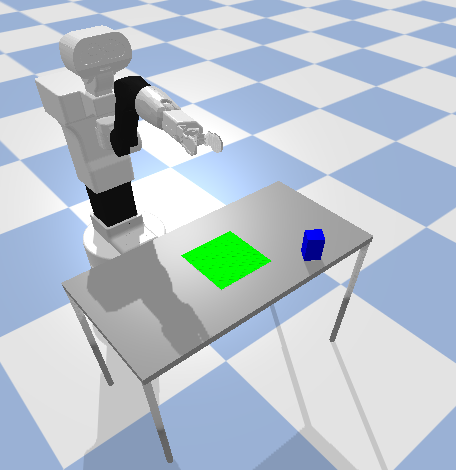}}
\caption{The block-pushing task involving one object. The objective is to ensure the blue block is in the green region.}
\label{fig:setup}
\end{figure}

We evaluate our method in a block push domain inside a physics simulator, as shown in Figure \ref{fig:setup}.
We use the PyBullet simulator for accurate collision simulation \cite{coumans2016pybullet}. 
The agent is a TIAGo mobile robot platform, which has a differential drive base, a prismatic torso joint, a 7 \ac{DoF} arm and a 2 \ac{DoF} parallel jaw gripper. The arm inverse and forward kinematics are computed using IKFast \cite{diankov_automated_2010}.

\subsection{Task Domain Description}
The task domain is almost identical to \textit{Domain 1} as implemented in the PDDLStreams library\footnote{\url{https://github.com/caelan/pddlstream}} and described in the paper \cite{garrett_pddlstream_2020}, in which several objects are placed on a table, and the goal is to place all the objects within a square region centered at the middle of the table. The placement is sampled from a uniform random distribution. The difference is that in our setup, some objects are specified as not \textit{graspable}. As the author of the library notes, the problem quickly becomes challenging as the number of objects increases, as the region becomes tightly packed with objects.

\subsection{Model Description}
Our \ac{PDDL} model builds on the model developed in the original PDDLStreams library.
Below we describe relevant components 
to accommodate our new task domain.




\textbf{Operators}
The robot is equipped with the primitive skills \texttt{move\textunderscore base}, \texttt{pick}, \texttt{place}, and \texttt{align}, and is tasked to learn the skill \texttt{push}, for which a scripted subroutine and symbolic operator are provided.
The skill \texttt{align} is a variant of \texttt{pick} in which the gripper is placed next to an object (instead of around it) so it can be pushed. The scripted subroutine for \texttt{push} simply moves the gripper toward the goal object position while keeping the gripper orientation constant.


The primitive skills are engineered as part of the PDDLStream library and inspired by an existing implementation of \texttt{move\textunderscore base}, \texttt{pick}, and \texttt{place} for a PR2 robot. Our implementation
is open-sourced as a fork of the original project.
The task plan is generated by \textit{FastDownward} \cite{helmert_fast_2006}.
Arm joint and base motion are planned using \textit{Bidirectional RRT} \cite{kuffner_rrt-connect_2000}.
The PDDLStream algorithm used is the \textit{Adaptive} algorithm.

\subsection{Algorithmic Details}

\textbf{Imitation Learning}
For the baseline policy, a variant of \ac{BC} that uses a Recurrent Neural Network (RNN) as the policy network was trained. The RNN allows the policy to model temporal dependencies through the recurrent hidden state.

\textbf{Offline RL}
For the \ac{ORL} policy, a variant of \ac{BCQ} is used.
\ac{BCQ} addresses the issue in offline \ac{RL} where target values are overestimated when querying the Q-network on actions unseen in the dataset, by approximately constraining the Q-network maximization to actions seen in the dataset. 

\textbf{Training Details}
All hyperparameters are taken from the default values chosen by Mandlekar \textit{et al.} \cite{mandlekar_what_2023} for the machine-generated dataset (c.f. Appendix D.3 of the article).
All policies were trained using the Robomimic library\footnote{\url{https://robomimic.github.io/}} on a machine equipped with a GeForce GTX1080Ti graphics card.

\subsection{Evaluation Protocol}
For all experiments, the success criterion is whether the goal description is satisfied at the end of the task execution; namely, whether all objects in the problem instance are contained within the square region placed in the middle of the table.
Note that problem instances that are not solved by the \ac{TAMP} component are discarded from the evaluation dataset because the aim of the experiments is not to evaluate the \ac{TAMP} solver.

%% file: results.tex
\section{Experimental Results and Discussion}

\subsection{\label{sec:exp1}Experiment 1: Robustness to sub-optimal demonstrations}

\begin{figure}[h]
    \centering
    \includegraphics[width=0.5\textwidth]{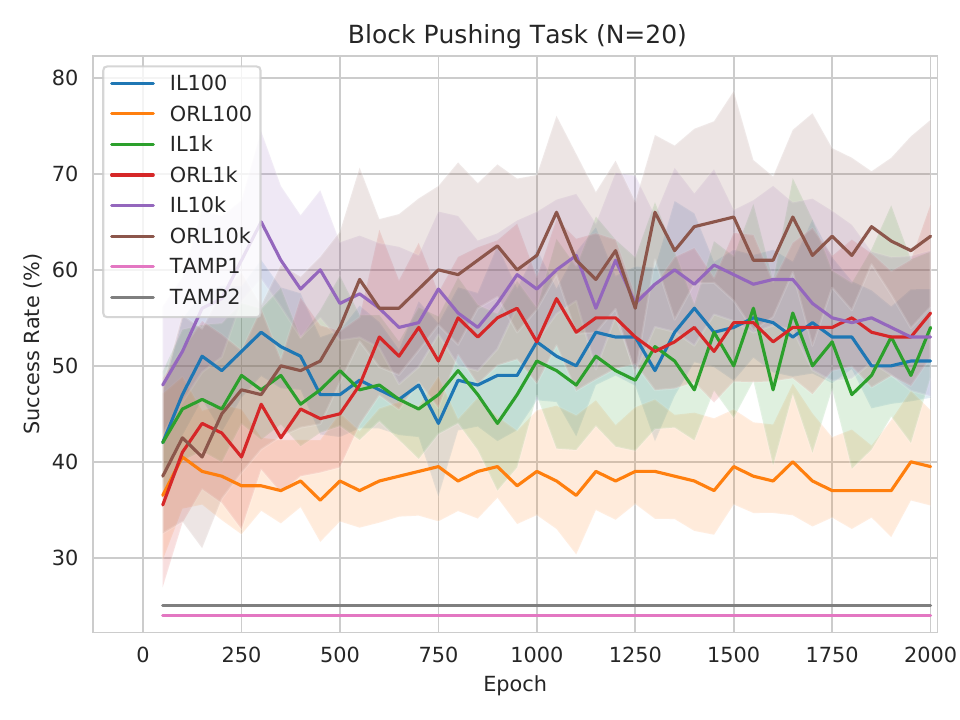}
    \caption{Success rates for \ac{IL} and \ac{ORL}-based policies trained on 100, 1000 and 1000 demonstrations over 2000 epochs vs. scripted subroutines on a pushing task involving one object (N=20). The lines denote the median and the shaded areas denote the 90\% bootstrapped confidence interval.}
    \label{fig:exp1}
\end{figure}

This experiment evaluates whether the suboptimal demonstrations provided by a scripted subroutine from a \ac{TAMP} solver can be used to train, in a fully offline manner, a reactive policy that outperforms the subroutine.
The environment is instantiated with one object for both training and evaluation. 
The success rate of the \ac{IL} and \ac{ORL} policies, trained on 100, 1000 and 10,000 demonstrations (of which 10\% is used for validation) over 2000 epochs (evaluated on checkpoints every 50 epochs) over 100 problem instances and averaged over 20 seeds are compared to the baseline scripted subroutine in Figure \ref{fig:exp1}. Note that policies trained on 10,000 demonstrations were evaluated on separate problem instances, hence the two TAMP baselines.

First, it is worth noting that the success rates of the baseline subroutines are just 24\% and 25\%.
All 6 learned policies outperform these baselines, showing generalization beyond the original dataset.
The fact that \ac{IL} can outperform the expert is most likely due to the expert relabeling procedure, which provides an optimal label for the goal-conditioned policies.
However, none of the \ac{IL} policies improve significantly with increasing training epochs.
Although there is no discernible performance gain of the \ac{IL} policy trained with 1000 demonstrations compared to 100, the performance increases with 10,000 demonstrations.
The \ac{ORL} policy with 100 demonstrations does not improve with more training epochs either, and performs worse than the \ac{IL} policies, suggesting that the dataset size is insufficient for generalization.
On the other hand, the \ac{ORL} policies with 1000 and 10,000 demonstrations do improve with more training epochs and outperforms the corresponding \ac{IL} policies after approximately 500 epochs. Notably, this policy consistently shows a two-fold improvement over the baseline subroutines beyond 500 epochs.
In summary, the experimental data indicates that given sub-optimal demonstrations of a pushing maneuver from a TAMP solver, \ac{ORL} with expert relabeling shows improved generalization over \ac{IL} given sufficient demonstration data.

\subsection{Experiment 2: Sample efficiency}

\begin{figure}[h]
    \centering
    \includegraphics[width=0.5\textwidth]{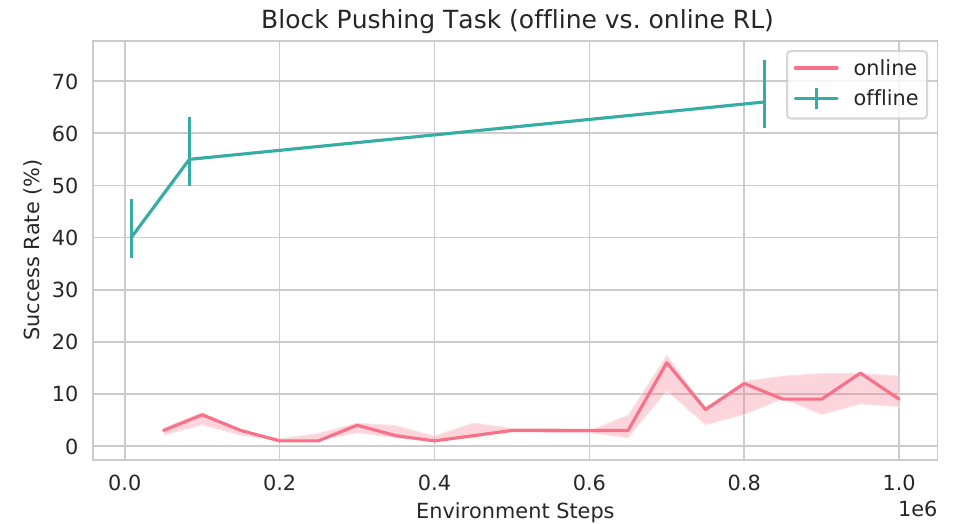}
    \caption{Success rates for online and offline RL policies for a blcok-pushing task involving one object. The lines denote the median, and the shaded area denotes the interquartile range.}
    \label{fig:exp2}
\end{figure}

This experiment aims to evaluate whether a policy trained fully offline with suboptimal TAMP demonstrations can outperform a policy trained online with less environment transitions.
The chosen online \ac{RL} algorithm is Truncated Quantile Critics (TQC) \cite{kuznetsov_controlling_2020}, equipped with a Hindsight Experience Replay Buffer \cite{andrychowicz_hindsight_2017}, and trained with a sparse reward given when the object position is within 5 cm of the goal, following the results from Plappert \textit{et al.} \cite{plappert_multi-goal_2018}. The hyperparameters are chosen from the benchmark of the same algorithm on RL Zoo \textit{FetchPush} environment, which also features a block-pushing task with a 7 \ac{DoF} manipulator arm, albeit with different observation and action spaces.
Namely, the action space consists of the arm joints and the observation space consists of the pose and velocities of the end-effector and the object.
The horizon is set to 100 steps, and the control frequency is set to 20 Hz.
To allow a fair comparison with the offline policy, at the start of every training episode, the initial environment configuration and the goal pose are sampled from the TAMP solver, such that the base and the gripper are well aligned with the block.
The success rate of the policy trained for 10,000 episodes (e.g. 1 million interactions), with checkpoints collected every 500 episodes, averaged over 3 seeds, is evaluated on 100 problem instances.
The results are compared to those of the offline policies ORL100 at 1950 epochs, ORL1k at 1800 epochs and ORL10k at 1300 epochs from Experiment 1. The trajectories in the training dataset total 8333, 83342, and 825770 environment transitions respectively, as shown in Figure \ref{fig:exp2}.

The online policy never reaches 24\%, which is the success rate of the TAMP baseline in Experiment 1.
While the algorithm hyperparameters were not tuned for this particular environment, the sensitivity of online \ac{RL} algorithms is highlighted by several works \cite{henderson_deep_2018, ibarz_how_2021}.
In addition to the large performance gap observed, it is important to note that the online interactions are not checked for collisions like the offline \ac{TAMP} demonstrations, making the online approach difficult to execute in physical setups.

\subsection{Experiment 3: Generalization to long-horizon tasks}

This experiment aims to evaluate whether the trained policy can be re-used in longer-horizon tasks than the ones it was trained on. The environment now consists of several objects, one of which is not graspable (i.e. needs to be pushed).
The task becomes more difficult with an increasing number of objects, as the unobstructed area of the goal region decreases. However, since the TAMP solver addresses the geometric reasoning aspect, and provides the initial and goal states for each push skill, we hypothesize that the push skill performance does not acutely degrade.

The environment is instantiated with 1, 2, and 3 objects. In the case of 3 objects, the shortest possible task plan consists of a sequence of 9 skills, consisting of \texttt{align}, \texttt{push}, \texttt{pick}, \texttt{place}, and \texttt{move\_base}.
The chosen \ac{IL} \& \ac{ORL} policies are IL1k at 1550 epochs and ORL1k at 1800 epochs from Experiment 1. As in Experiment 1, the results are averaged over 20 seeds. Table \ref{tab:multi} shows the success rates of each policy and the \ac{TAMP} subroutine over 100 problem instances.

\begin{table}[h]
\centering
\begin{tabulary}{\textwidth}{C|C C C}
\hline
\multirow{2}{*}{\textbf{\# objects}} & \multicolumn{3}{c}{\textbf{Success Rate (\%)}} \\
 & IL1000 & ORL1000 & TAMP \\\hline
1 & 56.0 $\pm$ [7.6, 10.9] & 55.0 $\pm$ [5.1, 8.1] & 24 \\
2 & 62.0 $\pm$ [7.0, 9.7] & 67.0 $\pm$ [4.7, 8.5] & 23 \\
3 & 54.0 $\pm$ [7.2, 9.8] & 57.5 $\pm$ [4.5, 7.2] & 18 \\
\end{tabulary}
\caption{Median push skill success rates ($\pm$90\% bootstrapped confidence intervals) for \ac{IL} and \ac{ORL}-based policies vs. subroutine for increasing number of objects.} 
\label{tab:multi}
\end{table}

Note that the first row is extracted directly from Experiment 1. 
As expected, the success rates of the learned policies do not degrade with more objects, in contrast to those of the subroutine. The rates improve in the 2-object case for both policies, while it drops 1\% for the subroutine. 
With 3 objects, however, the subroutine rate drops, and the policy rates return to approximately the same levels as in the 1-object task.
In summary, the data indicates that the modularity of \ac{TAMP} can be combined with the robustness of \ac{IL} / \ac{ORL}-based policies to generalize learned skills to long-horizon tasks without loss in performance relative to the baseline subroutine.

%% file: conclusion.tex
\section{Conclusion}
Generalizing robot manipulation skills for long-horizon tasks is difficult because it entails both abstract reasoning and robustness against environmental perturbations. 
Planning-based methods such as TAMP are not robust but can provide state and action abstraction, while learning-based methods can provide robustness given skill-dependent abstractions.
We introduce a novel skill-learning architecture that leverages suboptimal demonstrations from an integrated TAMP solver to train a goal-conditioned RL policy fully offline. Through simulated experiments, the learned skill was shown to outperform the suboptimal expert, be more sample efficient than a SOTA online \ac{RL} method, and be reusable in long horizon tasks, leading to more persistent robot learning. Future work includes expanding the skill repertoire, using image observations, and evaluating the approach on real hardware.